\documentclass{article}

\usepackage{microtype}
\usepackage{graphicx}
\usepackage{threeparttable}
\usepackage{subcaption}
\usepackage{booktabs} 
\usepackage{multirow}
\usepackage{makecell}
\usepackage{enumitem}
\usepackage{hyperref}
\usepackage{tcolorbox}
\tcbuselibrary{breakable} 
\newtcolorbox{promptbox}[1][]{
  colback=gray!5!white, 
  colframe=gray!75!black, 
  title={Prompt Example}, 
  fonttitle=\bfseries,
  breakable, 
  #1
}

\newcommand{\CRRM}{\textit{ConceptRM}}

\usepackage[preprint]{icml2026}

\usepackage{amsmath}
\usepackage{amssymb}
\usepackage{mathtools}
\usepackage{amsthm}

\newcommand{\M}[1]{{\color{black}#1}}

\usepackage[capitalize,noabbrev]{cleveref}

\theoremstyle{plain}

\theoremstyle{definition}

\theoremstyle{remark}

\usepackage[textsize=tiny]{todonotes}

\begin{document}

\twocolumn[
   \icmltitle{\CRRM: The Quest to Mitigate Alert Fatigue through Consensus-Based Purity-Driven Data Cleaning for Reflection Modelling}



  \icmlsetsymbol{equal}{*}

  \begin{icmlauthorlist}
    \icmlauthor{Yongda Yu}{equal,nju}
    \icmlauthor{Lei Zhang}{equal,nju}
    \icmlauthor{Xinxin Guo}{nju}
    \icmlauthor{Minghui Yu}{nju}
    \icmlauthor{Zhengqi Zhuang}{nju}
    \icmlauthor{Guoping Rong}{nju}
    \icmlauthor{Haifeng Shen}{scu}
    \icmlauthor{Zhengfeng Li}{alibaba}
    \icmlauthor{Boge Wang}{alibaba}
    \icmlauthor{Guoan Zhang}{alibaba}
    \icmlauthor{Bangyu Xiang}{alibaba}
    \icmlauthor{Xiaobin Xu}{alibaba}
  \end{icmlauthorlist}

  \icmlaffiliation{nju}{Software Insititute, Nanjing University, Nanjing, China}
  \icmlaffiliation{scu}{Southern Cross University, Gold Coast, Australia}
  \icmlaffiliation{alibaba}{TRE, Alibaba Inc., Hangzhou, China}

  \icmlcorrespondingauthor{Guoping Rong}{ronggp@nju.edu.cn}
  \icmlcorrespondingauthor{Zhengfeng Li}{lizhengfeng.lzf@alibaba-inc.com}
  \icmlkeywords{Machine Learning, ICML}

  \icmlkeywords{Code Review, Reflection Model}

  \vskip 0.3in
]



\printAffiliationsAndNotice{}  

\begin{abstract}
In many applications involving intelligent agents, the overwhelming volume of alerts (mostly false) generated by the agents may desensitize users and cause them to overlook critical issues, leading to the so-called ``alert fatigue''. A common strategy is to train a reflection model as a filter to intercept false alerts with labelled data collected from user verification feedback. However, a key challenge is the noisy nature of such data as it is often collected in production environments. As cleaning noise via manual annotation incurs high costs, this paper proposes a novel method \CRRM\ for constructing a high-quality corpus to train a reflection model capable of effectively intercepting false alerts. With only a small amount of expert annotations as anchors, \CRRM\ creates perturbed datasets with varying noise ratios and utilizes co-teaching to train multiple distinct models for collaborative learning. By analyzing the consensus decisions of these models, it effectively identifies reliable negative samples from a noisy dataset. Experimental results demonstrate that \CRRM\ significantly enhances the interception of false alerts with minimal annotation cost, outperforming several state-of-the-art LLM baselines by up to 53.31\% on in-domain datasets and 41.67\% on out-of-domain datasets.
\end{abstract}

\section{Introduction}
Recent times have witnessed the widespread adoption of intelligent agents in many applications. An example is code review, a practice that underpins modern software development~\cite{gousios2014exploratory}, where intelligent agents powered by Large Language Models (LLMs) are often used to automatically analyze code changes and generate review comments to assist human developers in identifying potential defects and suggesting improvements~\cite{peng2025icodereviewer, lu2023llama, yu2024fine, sun2025bitsai, ren2025hydra,li2025issue}. A common challenge faced by these applications is the so-called ``alert fatigue'' where an overwhelming volume of alerts (largely false positives) requires massive human verification to ensure true positives are not missed. Such a situation may desensitize users, causing them to overlook critical issues~\cite{gao2025systematic}. In the case of LLM-based code review, the agents may generate enormous low-quality comments, such as hallucinations or trivial nitpicks~\cite{sun2025bitsai, alansari2025large, agarwal2024codemirage}, potentially leading to ``alert fatigue"~\cite{gao2025systematic} and further causing developers to overlook critical vulnerabilities~\cite{sun2025bitsai}.

A common strategy to mitigate alert fatigue is to deploy a reflection model as a filter~\cite{madaan2023self} to intercept false alerts. Training such a reflection model requires high-quality labelled data comprising vast true positive and true negative samples~\cite{wang2024empirical}. However, in some applications, data is collected from production environments and often contains substantial noise, making it difficult to accurately identify true negatives~\cite{jiang2022softpatch}. In the case of code review, the data is sourced primarily from user feedback on LLM-generated review comments~\cite{cihan2025automated}. Due to various subjective and objective factors, developer feedback may not accurately reflect the actual quality of the review comments. For instance, under tight development schedules, developers tend to mark a majority of review comments as ``ignored'' despite that some of them should be accepted~\cite{soderberg2022understanding}. As such, constructing a high-quality reflection model training dataset often requires cleaning noise. Although data cleaning via manual annotation could address this issue, the associated workload and cost render it impractical. 

In this paper, we propose a novel method \CRRM\ (\underline{Co}nsensus-based \underline{N}oise \underline{C}l\underline{e}aning and \underline{P}urity-driven co-\underline{T}eaching for \underline{R}eflection \underline{M}odelling) for constructing a high-quality corpus to train a reflection model capable of effectively intercepting false alerts. Its key idea is to leverage a small amount of expert annotations as anchors to guide a consensus-based data cleaning strategy underpinned by the following two techniques.
\vspace{-0.1cm}
\begin{itemize}[topsep=0pt,itemsep=0pt,parsep=0pt,leftmargin=*]
  \item \emph{Noise-doping-based data cleaning}: Leveraging the property that neural networks prioritize learning simple, generalizable patterns (the ``memorization effect"), we inject ambiguous data (e.g., the ``ignored'' class in code review) within positive examples (e.g., the ``accept'' class in code review) at varying ratios. This perturbation shifts the models' decision boundaries, enabling the training of multiple diverse sub-models specifically for noise filtration.

\item \emph{Consensus-based biased relabeling}: We employ a biased selection objective to identify the optimal combination of sub-models for noise cleaning. By utilizing this consensus mechanism to relabel the training corpus, we effectively inject the specific optimization bias (e.g., strict constraints on false positives) directly into the training data.
\end{itemize}

Experimental results demonstrate that \CRRM\ effectively mitigates label noise, achieving a reduction in the false positive rate of up to 75.51\% compared to Naive fine-tuning. Furthermore, it significantly enhances the efficacy of intercepting false alarms, outperforming several state-of-the-art (SOTA) LLM baselines by up to 53.31\% with minimal annotation costs.

\section{Related Work}
In this section, we review related work in two areas: model hallucination and alert fatigue in LLM-based code review and Learning with Noisy Labels (LNL).

\subsection{Model Hallucination and Alert Fatigue in LLM-based Code Review} 
The advent of LLMs, with their profound capability to bridge natural language and code semantics, has catalyzed a paradigm shift in code review. While LLMs offer unprecedented capabilities, they may introduce a critical bottleneck: hallucination. In the context of code review, this phenomenon manifests in typical erroneous forms such as fabricated issue descriptions, phantom reference recommendations and erroneous fix suggestions~\cite{gao2025systematic}.

In addition, LLM hallucinations are often delivered in a persuasive and authoritative tone~\cite{zhou2025large}. This creates a high risk where junior developers may blindly accept erroneous suggestions, while senior developers are forced to expend significant effort to ``falsify'' them. Furthermore, the proliferation of low-quality or trivial review comments also exacerbates the verification cost, leading to the so-called ``alert fatigue and wasting developers' time~\cite{goldman2025types, vijayvergiya2024ai}. Crucially, due to the inherent overconfidence of LLMs, standard self-reflection mechanisms often fail to detect these hallucinations in a single turn~\cite{cheninside, zhang2025understanding}. Consequently, there is an urgent need to construct a robust reflection model specifically designed to assess the quality of comments generated by various LLMs.

\subsection{Learning with Noisy Labels}
The primary goal of Learning with Noisy Labels (LNL) is to develop robust algorithms that can converge to optimal solutions comparable to those trained on clean data, despite the presence of corrupted labels~\cite{han2020survey}. LNL approaches typically rely on the ``memorization effect" of Deep Neural Networks (DNNs), which posits that DNNs prioritize learning simple, generalizable patterns in the early training stages before overfitting to noisy samples or outliers later on~\cite{arpit2017closer}. Leveraging this property, Co-teaching~\cite{han2018co} employs two networks that cross-supervise each other by selecting small-loss samples, achieving remarkable denoising performance without prior knowledge of the noise distribution. Co-teaching+~\cite{yu2019does} further improves this by explicitly maintaining divergence between the two networks.

However, the prevailing paradigm shift from ``training from scratch" to ``re-training \& fine-tuning" poses unprecedented challenges to these classical LNL methods. Unlike training initialized with random weights, modern approaches fine-tune models starting from identical, high-quality weights pre-trained on massive corpora. This uniformity fundamentally undermines the initialization divergence upon which Co-teaching algorithms rely, causing the two networks to converge to similar states prematurely. To address this limitation, we propose shifting the source of divergence from the \textit{parameter space} to the \textit{data space}. 

By introducing data perturbation, we dynamically reconstruct view differences between models. We then utilize annotated data as anchors to select model combinations with high interception efficacy, use this combination to relabel the data, and finally train a robust interception model with high efficacy using the relabeled data.

\section{Task Definition}
\label{sec:task}
In the context of code review, the primary objective of a reflection model is to develop an automated mechanism capable of 
determining whether an LLM-generated review comment should be retained or intercepted based on its relevance to the given code change.

Let $C$ denote a given code \textit{diff} and $R$ represent the review comment generated by the LLM based on $C$. Our task is to learn a mapping function $f_{\theta}: (C, R) \to \{0, 1\}$. Here, $y=1$ indicates that the review comment is of ``High Quality" (and should be retained), while $y=0$ indicates ``Low Quality" (and should be intercepted). The following constraints should be noted:
\vspace{-.1cm}
\begin{itemize}[topsep=0pt,itemsep=0pt,parsep=0pt,leftmargin=*]
    \item First, the definition of low-quality comments ($y=0$) extends beyond mere hallucinations or factual errors to include High-Cost/Low-Reward 
    considerations. Specifically, if the remediation cost $\mbox{\it Cost(R)}$ associated with a suggestion $R$ significantly outweighs its potential gain $\mbox{\it Gain(R)}$—such as in cases of trivial naming conventions or excessive refactoring—the suggestion is classified as a negative sample.
    \item Second, 
    traditional optimization objectives (e.g., Accuracy or F1-Score) assume symmetric costs for misclassifications~\cite{elkan2001foundations}. However, in the case of code review, this assumption does not hold because the cost of a false positive (intercepting a high-quality review) is significantly higher than that of a false negative (failing to intercept a low-quality review). Erroneously intercepting a valuable suggestion may allow critical code defects to flow into the production environment, thereby degrading software quality~\cite{mcintosh2014impact}. 
\end{itemize}

Formally, we define the following two key metrics to measure the interception performance on the validation set $\mathcal{D}_{val}$:
\vspace{-.5cm}
\begin{itemize}[topsep=0pt,itemsep=0pt,parsep=0pt,leftmargin=*]
    \item Intercept Rate ($\mbox{\it IR}$): The proportion of low-quality comments successfully identified by the model (equivalent to Recall on the Negative Class).
    \item False Positive Rate ($\mbox{\it FPR}$): The proportion of high-quality comments erroneously intercepted by the model.
\end{itemize}

So we formulate this task as a multi-objective optimization problem, aiming to train a model $f_{\theta}$ that maximizes the Signal-to-Noise Gain: pursuing the maximal ratio of $\mbox{\it IR}$ to $\mbox{\it FPR}$, subject to a minimum coverage constraint. The optimization objective $\mathcal{L}_{obj}$ is defined as follows:
\begin{equation}
\max_{\theta} \Pi(f_{\theta}, \mathcal{D}_{val}) = \frac{\textit{IR}(f_{\theta})}{\textit{PFR}(f_{\theta}) + \epsilon}
\label{eq:pie}
\quad
\text{s.t.} \quad \textit{IR}(f_{\theta}) \geq \gamma
\end{equation}
where $\Pi$ represents what we call Purity-driven Intercept Efficacy (PIE), $\epsilon$ is a smoothing term, and $\gamma$ denotes the Minimum Coverage Threshold. This definition underscores our priority: rather than simply maximizing the absolute volume of intercepts, we seek to maximize the intercept yield per 1\% of false positive cost incurred, provided that the baseline intercept capability ($\gamma$) is met. \textbf{The purity objective indicates that the model should intercept only false alerts.} A full list of notation definitions is in Table \ref{tab:notation}.

\begin{table}[h]
    \centering
    \scriptsize
    \caption{Summary of Notations and Definitions}
    \label{tab:notation}
    \begin{tabular}{ll}
        \toprule
        \textbf{Notation} & \textbf{Description} \\
        \midrule
        $C$ & The code diff context submitted for review \\
        $R$ & The review comment generated by the LLM based on $C$ \\
        $f_{\theta}$ & The Reflection Model (mapping function) with parameters $\theta$ \\
        $y$ & Ground truth label ($1$: High Quality, $0$: Low Quality/Intercept) \\
        $\mbox{\it Cost(R)}$ & The remediation cost associated with suggestion $R$ \\
        $\mbox{\it Gain(R)}$ & The potential benefit gained from applying suggestion $R$ \\
        $\mathcal{D}_{\mbox{\it raw}}$ & The raw uncleaned dataset \\
        $\mathcal{D}_{\mbox{\it val}}$ & The validation set \\
        $\mbox{\it PIE}$ & Purity-driven Intercept Efficacy defined as $\mbox{\it IR} / (\mbox{\it FPR} + \epsilon)$ \\
        $\gamma$ & Minimum Coverage Threshold for interception \\
        $\epsilon$ & A smoothing term to prevent division by zero \\
        $\mathcal{S}$ & The set of perturbed training datasets $\{\mathcal{D}_1, \dots, \mathcal{D}_K\}$ \\
        $\alpha_k$ & The mixture ratio of ``ignore'' samples in the positive class for $\mathcal{D}_k$ \\
        $\mathcal{M}$ & The set of parallel models $\{M_1, \dots, M_K\}$ trained on $\mathcal{S}$ \\
        $\mathcal{C}$ & The consensus strategy, e.g., Strict Consensus (SC) or Majority Vote(MV) \\
        \bottomrule
    \end{tabular}
\end{table}

\begin{figure*}[!t]
    \centering                                      
    \includegraphics[width=0.86\linewidth]{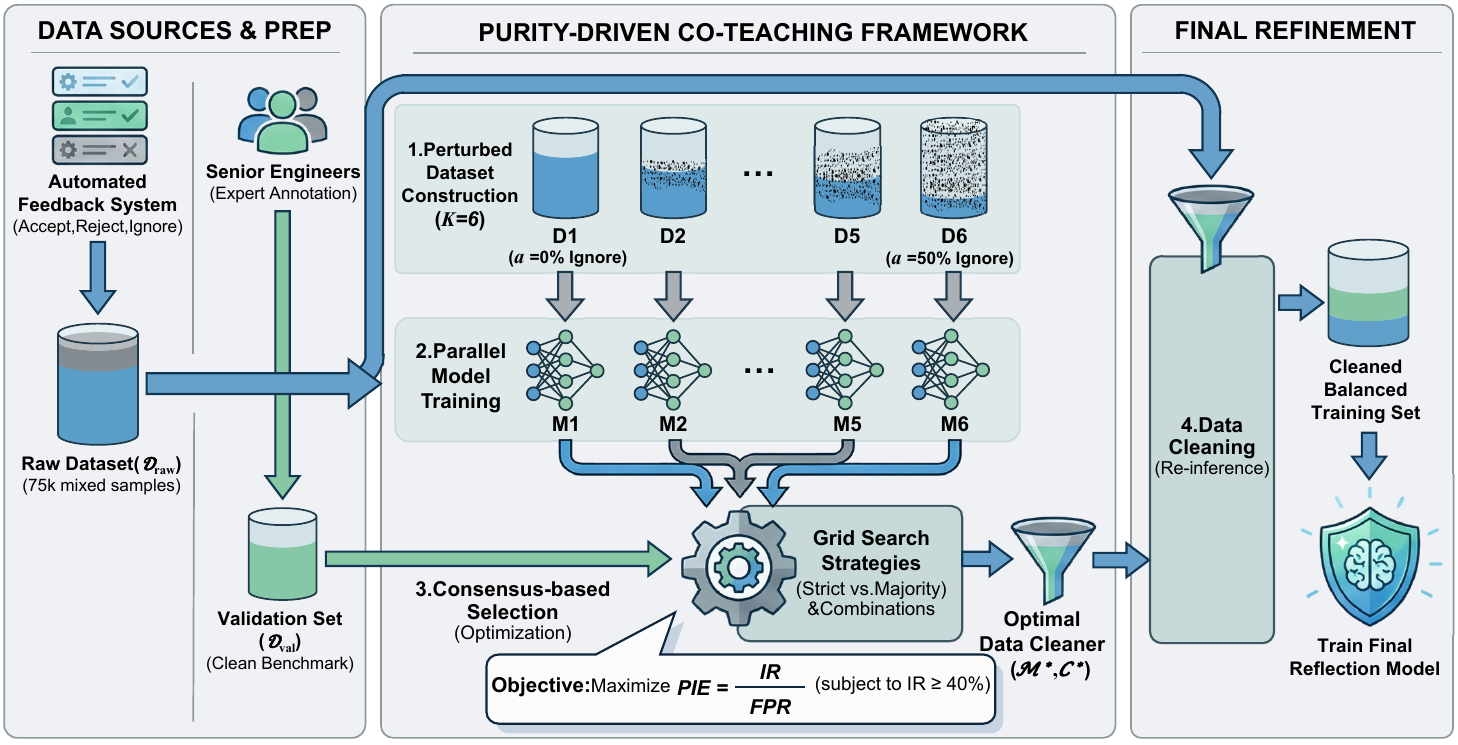}
    \caption{A schematic overview of \CRRM.}
    \label{fig:overview}
\end{figure*}

\section{Methodology}
Figure~\ref{fig:overview} presents an overview of the proposed \CRRM\ framework for training data preparation and reflection model training. Instead of relying on model initialization differences, our method constructs training sets with differentiated noise distributions. By utilizing a small set of expert-annotated data as anchors, we establish a consensus mechanism based on data perturbation to filter high-signal-to-noise samples from massive noisy data. This consensus strategy is explicitly guided to maximize the ratio of $\mbox{\it IR}$ to $\mbox{\it FPR}$, prioritizing safe interception over aggressive filtering.

\subsection{Data Acquisition and Pre-processing}
\label{sec:data_construction}
\paragraph{Data Source and Feedback Mechanism.}
The raw data was collected from the internal code review system of a global Internet company. The system operates on an agent-based architecture: it decomposes code commits, identifies potential code issues, 
and employs an intelligent agent to retrieve necessary context before outputting the final issues. A single code \emph{diff} may trigger multiple review comments. And users often label feedback with three options:
\vspace{-.1cm}
\begin{itemize}[topsep=0pt,itemsep=0pt,parsep=0pt,leftmargin=*]
    \item Accept: The user agrees with the suggestion (labeled as positive).
    \item Reject: The user considers the suggestion erroneous or valueless (labeled as negative).
    \item Ignore: The user takes no explicit action, often collapsing the comment view.
\end{itemize}
\vspace{-.1cm}
\paragraph{Expert Annotation.}
To establish a reliable evaluation benchmark, we collaborated with senior software engineers (5+ years of working experience) in the company to construct a validation set $\mathcal{D}_{val}$. Each data instance was independently annotated by three engineers, with the final label determined by majority vote. Annotation continued until a balanced dataset of 100 samples (50 positives and 50 negatives) was collected. 
In particular, if an issue raised by a review comment $R$ did not pose security or functional risks and satisfied $\mbox{\it Cost(R)} \leq \mbox{\it Gain(R)}$, it was labeled as a negative sample. 

\vspace{-.3cm}
\paragraph{Dataset Construction.}

We performed stratified sampling on all raw data with a ratio of Accept:Reject:Ignore = 1:1:1 to construct a raw, uncleaned dataset $\mathcal{D}_{raw}$.

\subsection{Noise-Robust Training via Purity-Driven Co-Teaching}
Traditional Co-Teaching methods filter noise by training two networks with different initial states and exchanging ``small-loss" samples~\cite{han2018co}. However, in the era of Pre-trained Language Models such as the pervasive LLMs adopted in code review, downstream models share identical pre-trained weights. Forcing divergent random initializations can lead to catastrophic forgetting of pre-trained knowledge. To address this, we designed a screening mechanism based on \textbf{Decision Boundary Perturbation}~\cite{miyato2018virtual, laine2017temporal}. Rather than relying on initialization variance, we perturbed the model's decision boundary by introducing noise data with varying distributions, using the PIE metric $\Pi$ defined Equation~\ref{eq:pie} to guide the final model fusion.

\subsection{Construction of Perturbed Datasets}
To create decision boundaries with varying preferences, we treated the highly ambiguous ``Ignore" data as a noise source. We construct $K=6$ training sets $\mathcal{S} = \{\mathcal{D}_1, ..., \mathcal{D}_K\}$ with different noise mixture ratios. For each dataset $\mathcal{D}_k$, we maintained a balance of $N_{pos}:N_{neg} = 1:1$. The negative samples were explicit ``Reject'' data. The positive samples, however, were composed of ``Accept'' data mixed with ``Ignore'' data at a ratio $\alpha_k$ (incrementing from $0\%$ to $50\%$ with a 10\% step size). This construction forced the model $M_k$ trained on $\mathcal{D}_k$ to learn a distinct decision boundary:
\vspace{-.1cm}
\begin{itemize}[topsep=0pt,itemsep=0pt,parsep=0pt,leftmargin=*]
    \item Conservative high-noise models, i.e., low $\mbox{\it FPR}$, low $\mbox{\it IR}$.
    \item Aggressive low-noise models, i.e., high $\mbox{\it FPR}$, high $\mbox{\it IR}$.
\end{itemize}

The rationale is that although the positive-negative ratio is strictly balanced (1:1), injecting ``Ignore" samples into the positive class causes a significant semantic distribution shift. ``Ignore" instances conceptually occupy a grey area between acceptance and rejection, sharing latent features (e.g., triviality, verbosity) with the ``Reject" class. In a high-noise environment, the model is penalized for classifying ambiguous samples as negatives. Consequently, the optimization process pushes the decision boundary deep into the negative region to accommodate these ``hard positives". This forces the model to adopt a conservative strategy, flagging only the most egregious errors as negatives to minimize $\mbox{\it FPR}$, which inevitably lowers the $\mbox{\it IR}$. Conversely, models trained on cleaner data adopt stricter quality standards, resulting in broader interception but higher misclassification rates. We utilize this divergence in decision boundaries, combined with the expert-annotated anchor data, to extrapolate the optimal decision plane. By treating ambiguous ``Ignore" samples as positives, we explicitly bias the model to classify uncertain boundary cases as ``Accept''. This forces the decision boundary deep into the negative territory, ensuring that only high-confidence defects are intercepted.

\subsection{Consensus-based Selection for Optimal Model Combination}
Unlike traditional methods that blindly ensemble all models, we aimed to identify the optimal combination that maximizes purity-driven intercept efficacy $\Pi$ (Equation~\ref{eq:pie}). We trained the model set $\mathcal{M} = \{M_1, ..., M_K\}$ in parallel and periodically save checkpoints. During the validation phase, we introduced a Consensus Strategy $\mathcal{C}$ to fuse model predictions. We compare two strategies, i.e., Strict Consensus (SC) vs. Majority Vote (MV), as below:
\vspace{-.1cm}
\begin{itemize}[topsep=0pt,itemsep=1pt,parsep=0pt,leftmargin=*]
    \item SC: $\hat{y} = \bigwedge_{m \in \mathcal{M}_{\mbox{\it sub}}} (\mbox{\it m(x) ==} \text{Negative})$
    \item MV: $\hat{y} = \text{Mode}_{m \in \mathcal{M}_{sub}}(m(x))$
\end{itemize}
In these formulations, $x$ denotes the input sample and $\hat{y}$ represents the final ensemble prediction. The term $m(x)$ refers to the output label predicted by an individual model $m$ within the selected subset $\mathcal{M}_{\textit{sub}} \subseteq \mathcal{M}$. Specifically, the symbol $\bigwedge$ denotes the logical conjunction (AND), implying that SC classifies a sample as negative only if there is unanimous agreement among all models in $\mathcal{M}_{\textit{sub}}$, thereby prioritizing a low $\textit{FPR}$. In contrast, the $\text{Mode}$ operator in MV assigns the label based on the majority frequency, balancing $\textit{IR}$ and $\textit{FPR}$.

To select the optimal ``data cleaner" from the numerous model combinations and strategies, we performed a grid search on $\mathcal{D}_{val}$ based on the optimization objective defined in the Task Definition. Specifically, we selected the best combination $(\mathcal{M}^*, \mathcal{C}^*)$ that satisfies:
\vspace{-.2cm}
\begin{equation}
\label{eq:grid}
\small
\begin{split}
(\mathcal{M}^*, \mathcal{C}^*) = \operatorname*{argmax}_{\mathcal{M}', \mathcal{C}'} \left( \frac{IR(\mathcal{M}', \mathcal{C}')}{\mbox{\it FPR}(\mathcal{M}', \mathcal{C}')} \right) \quad\\
\text{subject to} \quad \mbox{\it IR}(\mathcal{M}', \mathcal{C}') \geq \gamma
\end{split}
\end{equation}

In this formulation, $(\mathcal{M}^*, \mathcal{C}^*)$ denotes the optimal subset of models and consensus strategy identified by the search. The objective function maximizes the ratio of $\textit{IR}$ to $\textit{FPR}$. The constraint ensures that the solution meets a minimum utility requirement, where $\gamma$ represents the lower bound threshold for $\textit{IR}$. We set $\gamma=40\%$ to maintain a baseline interception capability while prioritizing models that effectively filter low-quality comments.

\subsection{Final Refinement}
Using the selected optimal combination $(\mathcal{M}^*, \mathcal{C}^*)$, we performed re-inference (i.e., relabel) on the full raw dataset (containing all ``Accept'', ``Reject'', and ``Ignore'' samples). We constructed a final balanced training set using this cleaned data to train the ultimate Reflection Model. Essentially, this process made the implicit patterns within the ``Ignore" data explicit to complete the denoising process with the objective to maximize the purity-driven intercept efficacy $\Pi$ (Equation~\ref{eq:pie}).

\section{Experiments}
In this section, we present a comprehensive experimental evaluation to validate the effectiveness and robustness of the \CRRM\ method in intercepting false alerts (low-quality comments) in LLM-based code review. 

\subsection{Experimental Design}
The primary objective of the experiments is to verify whether \CRRM\ can: (1) effectively filter out noise from industrial feedback data, and (2) establish a high-precision interception boundary that minimizes interference with developers to mitigate alert fatigue. Specifically, our experimental design aims to answer the following questions:
\vspace{-.3cm}
\begin{itemize}[topsep=0pt,itemsep=0pt,parsep=0pt,leftmargin=*]
\item \textbf{Performance Verification:} How does \CRRM\ perform against naive fine-tuning strategies and state-of-the-art LLMs in terms of interception accuracy and false positive control?
\item \textbf{Factor Analysis:} How do critical factors—such as the volume of expert annotation, consensus strategies, and label cleaning mechanisms—impact the reflection model's performance?
\end{itemize}
To systematically address these questions, we constructed both proprietary industrial datasets and open-source benchmarks, employed strict evaluation metrics, and compared our approach against a diverse set of baselines. The overall experimental design is summarized in Table~\ref{tab:experiment_design}.
\vspace{-.2cm}
\begin{table}[bpht!]
\caption{Overview of the Experimental Design.}
\label{tab:experiment_design}
\begin{center}
\scriptsize
\begin{tabular}{p{0.05\textwidth} p{0.24\textwidth} p{0.13\textwidth}}
\toprule
\textbf{Component} & \textbf{Description \& Method} & \textbf{Verification Objective} \\
\midrule
\textbf{Datasets} & \textbf{Internal Set:} Real-world industrial workflow (Train/Val/Test). \newline \textbf{Open Source Set:} Top 10 languages from Github (Generalizability Test). & Ensure robustness across domain-specific and general scenarios. \\
\hline
\textbf{Baselines} & \textbf{Naive Fine-tuning:} Training on raw, noisy data. \newline \textbf{SOTA LLMs:} Prompt Engineering. & Benchmark against noise-intolerant methods and flagship models. \\
\hline
\textbf{Metrics} & \textbf{IR:} Recall of low-quality comments. \newline \textbf{FPR:} Error rate of high-quality comments. \newline \textbf{PIE:} Consolidated purity-driven intercept efficacy ($\mbox{\it IR/FPR}$). & Quantify the trade-off between interception capability and user disturbance. \\
\hline
\textbf{Ablation} & \textbf{Data Volume:} 100 vs. 1,000 expert annotations. \newline \textbf{Consensus Strategy:} Strict Consensus (SC) vs. Majority Vote (MV). & Analyze data efficiency and decision boundaries. \\
\bottomrule
\end{tabular}
\end{center}
\end{table}
\vspace{-.7cm}

\subsection{Experimental Setup}
\paragraph{Internal Training Set.}This dataset corresponds to the uncleaned dataset $\mathcal{D}_{raw}$ described in Section~\ref{sec:data_construction}, comprising a total of 74,876 samples. The training set strictly follows the stratified sampling strategy (1:1:1 ratio) mentioned above.

\paragraph{Internal Validation Set.}
Following the Expert Annotation protocol established in Section~\ref{sec:data_construction}, we expanded the validation set $\mathcal{D}_{val}$ from the initial pilot scale (100 samples) to a total of 1,000 samples to further investigate the impact of dataset size on performance. The final validation set maintains a balanced 1:1 ratio between positive and negative instances, with ground truth determined by the majority vote of domain experts.

\paragraph{Internal Test Set.} The internal test set was acquired and annotated using the identical protocol applied to the internal validation set. It comprises 1,000 samples with a balanced 1:1 positive-to-negative ratio. The training, validation, and test sets are mutually exclusive, ensuring no data leakage.

\paragraph{Open Source Test Set.} The open-source community test set was constructed based on the top ten most popular programming languages identified in the Stack Overflow Developer Survey~\cite{stackoverflow2025survey}, i.e, JavaScript, Python, TypeScript, Java, C\#, C++, C, PHP, Go, and Rust. For each language, we selected the top five repositories by star count between December 2024 and December 2025, excluding repositories with fewer than 1,000 Pull Requests (PRs). From each repository, we extracted PRs that were successfully merged into the main branch and contained multiple human review comments, totalling 200 PRs.
We utilized the company's internal Agent framework to replay these PRs and generate review comments. To ensure the generalizability of the data distribution, we employed seven distinct models—encompassing both open-source and proprietary flagship models—to generate the initial comments. 
All generated comment data was shuffled and annotated by the same cohort of senior engineers using a consensus mechanism analogous to that of the internal dataset. The annotation process continued until an open-source test set of 1,000 samples with a balanced 1:1 positive-to-negative ratio was established.
\vspace{-.2cm}
\paragraph{Evaluation Metrics.}
We use $\mbox{\it IR}$ to evaluate the model's ability to correctly intercept low-quality comments and $\mbox{\it FPR}$ to evaluate its error rate in mistakenly intercepting high-quality comments.
\vspace{-.2cm}
\begin{align*}
\mbox{\it IR} = \frac{\mbox{\it TP}}{\mbox{\it TP + FN}} \qquad
\mbox{\it FPR} = \frac{\mbox{\it FP}}{\mbox{\it FP + TN}}
\end{align*}
where $\mbox{\it TP}$ represents correctly intercepted low-quality comments, $\mbox{\it FP}$ represents incorrectly intercepted high-quality comments, $\mbox{\it FN}$ represents missed low-quality comments (false negatives), and $\mbox{\it TN}$ represents correctly passed high-quality comments.

Inspired by the concept of Signal-to-Noise Ratio (SNR)~\cite{hoult1976signal}, 
we introduce $\mbox{\it PIE}$ (Purity-driven Intercept Efficacy) as a consolidated metric to quantify the interception yield per unit of false positive cost, aiming to identify the optimal decision boundary that correctly filters out low-quality comments without incorrectly discarding high-quality ones. $\mbox{\it PIE}$ is 
formally defined as:
\vspace{-.2cm}
$$\mbox{\it PIE} = \mbox{\it IR}/{\mbox{\it FPR}}$$

\paragraph{Baselines.}
The baselines consist of two categories. The first category is training-based, which utilizes standard training without using our proposed data cleaning strategy. It serves to demonstrate the ablation improvement of our framework. To further evaluate performance under low-resource conditions, we introduce a validator consisting of 100 data samples (50 positive and 50 negative) randomly extracted from the validation set to guide the construction of the data cleaner. The data cleaner employs a Majority Voting (MV) consensus method.

The second category is Prompt Engineering-based, which evaluates the performance of state-of-the-art LLMs in intercepting low-quality comments via prompt engineering. The models include Qwen-30B-A3B-Instruct-2507, Qwen-30B-A3B, Qwen-235B-A22B, Qwen-235B-A22B-Instruct-2507, GLM-4.7, Deepseek-V3.2, Kimi-K2-Instruct-0905, Qwen-Coder-480B, and Claude-4.5-Sonnet. We selected the latest flagship open-source models from mainstream providers. Additionally, Qwen-30B-A3B was chosen to facilitate a pre and post-training comparison, while Claude-4.5-Sonnet was included to benchmark the performance of commercial models. For models supporting mixed reasoning strategies, we evaluate both ``thinking" and ``no-think" modes. Due to data privacy constraints preventing the use of external models on internal data, we primarily use open-source models for comparison on the internal dataset, while Claude-4.5-Sonnet is evaluated exclusively on the GitHub (open-source) dataset.
\vspace{-.2cm}
\begin{table}[H]
\caption{Training hyperparameters.}
\begin{center}
\scriptsize
\begin{tabular}{|c|c|c|c|c|c|}
\hline
epochs & batch size & learning rate & cutoff & warmup steps & optimizer \\
\hline
2 & 128 & 1e-6 & 20000 & 100 & AdamW \\
\hline
\end{tabular}
\end{center}
\label{tab:TrainingHyp}
\end{table}
\vspace{-0.8cm}
\paragraph{Implementation Details.}
For data cleaning and final model training, we utilize Qwen-30B-A3B-Instruct-2507 as the base model. This is because Qwen-30B-A3B-Instruct-2507 strikes an optimal balance between a large total parameter count and a smaller number of active inference parameters, effectively harmonizing performance with efficiency. The training hyperparameters are detailed in Table~\ref{tab:TrainingHyp}.
For evaluations using prompt engineering, we apply the optimal parameter combinations recommended for each specific model. When evaluating our proposed model, we set the temperature to 0.2 to minimize inference randomness.

\section{Results and Analysis}
The analysis of results is divided into two parts. First, we verify \CRRM's overall performance compared to the baselines. Second, we analyze the internal factors and mechanisms that contribute to its superior performance.

\begin{table}[]
\begin{threeparttable}
\centering
\setlength{\tabcolsep}{3.5pt} 
\caption{Interception Performance: \CRRM\ vs. Baselines. \M{The PIE value of 4.42 from \CRRM-SC-1k translates to an average of 4.42 low-quality comments being intercepted per every high-quality comment mis-intercepted, which has a higher interception efficacy than Claude-4.5-Sonnet's 1.53. }}
\label{tab:codereview_refinement_performance}
\scriptsize
\begin{tabular}{lcccccc}
\toprule
\multirow{2}{*}{\textbf{Method}} & 
\multicolumn{3}{c}{\textbf{Inner Testset}} & 
\multicolumn{3}{c}{\textbf{Github Testset}} \\
\cmidrule(lr){2-4} \cmidrule(lr){5-7} 
& \textbf{IR} $\uparrow$ & \textbf{FPR} $\downarrow$ & \textbf{PIE}$\uparrow$ & \textbf{IR}$\uparrow$ & \textbf{FPR} $\downarrow$ & \textbf{PIE}$\uparrow$ \\
\midrule
QW30A3 & 34.6 & 19.4 & 1.78 & 33.0 & 12.6 & 2.62 \\
QW30A3T & 28.4 & 20.4 & 1.39 & 27.4 & 12.0 & 2.28 \\
QW30A3I & 31.0 & 12.8 & 2.42 & 30.6 & 9.8 & 3.12  \\
QW235A22 & 26.2 & 13.2 & 1.98 & 26.8 & 8.6 & 3.11 \\
QW235A22T & 25.6 & 11.2 & 2.29 & 25.2 & 8.8 & 2.86 \\
QW235A22I & 46.4 & 22.8 & 2.04 & 46.6 & 22.8 & 2.04 \\
QWCoder & 60.8 & 32.8 & 1.85 & 47.2 & 26.6 & 1.77 \\
DeepSeek-V3.2 & 52.2 & 32.2 & 1.62 & 30.4 & 36.8 & 0.83 \\
KimiK2 & 52.2 & 27.0 & 1.93 & 43.2 & 21.8 & 1.98 \\
GLM-4.7 & 57.6 & 26.0 & 2.21 & 42.0 & 20.4 & 2.05 \\
Claude-4.5-Sonnet & - & - & - & 71.0 & 46.4 & 1.53 \\
\hline
Naive Fine-Tuning & 73.0 & 34.8 & \makecell{2.10\\(-13.22\%)} & 78.0 & 39.2 & \makecell{1.99 \\(-36.22\%)} \\
\CRRM-SC-100 & 50.8 & 15.8 & \makecell{3.21 \\(+32.64\%)} & 48.6 & 13.8 & \makecell{3.52 \\(+12.82\%)} \\
\CRRM-MV-100 & 51.2 & 16.8 & \makecell{3.05 \\(+26.03\%)} & 51.8 & 16.2 & \makecell{3.20 \\(+2.56\%)} \\
\hline
\CRRM-SC-1k & 48.2 & 13.0 & \makecell{\textbf{3.71} \\(+53.31\%)} & 42.4 & 9.6 & \makecell{\textbf{4.42} \\(+41.67\%)}  \\
\CRRM-MV-1k & 56.6 & 18.8 & \makecell{3.01 \\(+24.38\%)} & 56.8 & 18.0 & \makecell{3.15 \\(+0.96\%)} \\

\bottomrule
\end{tabular}
\begin{tablenotes}
\scriptsize
\item QW30A3: Qwen3-30B-A3B (no think)~~QW30A3T: Qwen3-30B-A3B (thinking)
\item QW30A3I: Qwen3-30B-A3B-Instruct-2507
\item QW235A22: Qwen3-235B-A22B (no think)
\item QW235A22T: Qwen3-235B-A22B (thinking)
\item QW235A22I: Qwen3-235B-A22B-Instruct-2507
\item QWCoder: Qwen3-Coder-480B-Instruct~~~~KimiK2: Kimi-K2-Instruct-0905
\end{tablenotes}
\end{threeparttable}
\end{table}

\vspace{-.3cm}
\subsection{Performance Verification of \CRRM}
\paragraph{Comparison with Naive Fine-Tuning Baselines.} As shown in Table~\ref{tab:codereview_refinement_performance}, directly fine-tuning with uncleaned, noisy data yields relatively high IR on both test sets (Inner: 73.0\%, Github: 78.0\%). However, this is accompanied by high FPR (Inner: 34.8\%, Github: 39.2\%). This implies that for every five low-quality comments intercepted, two high-quality comments are erroneously filtered out. Such a high FPR is intolerable in real industrial deployment, as it significantly increases the probability of blocking valid feedback and raises the risk of overlooking critical code vulnerabilities.

In contrast, \CRRM\ methods significantly reduce FPR. Specifically, when employing the SC strategy, the FPR of \CRRM-SC-100 drops to 15.8\% on the Inner Testset and 13.8\% on the Github Testset, respectively. This improvement is evidently reflected by the $\mbox{\it PIE}$ metric, with \CRRM-SC-1k achieving the highest scores on both datasets (3.71 and 4.42, respectively), significantly outperforming the Naive fine-tuned models (2.10 and 1.99). This provides strong evidence that \CRRM\ effectively identifies and eliminates label noise in the training data, guiding the model to learn more robust classification boundaries. Consequently, it preserves valuable review comments while maintaining effective interception capabilities.

\vspace{-.2cm}
\paragraph{Comparison with Mainstream LLMs.}
We compare our lightweight model (based on Qwen-30B) with current flagship models using Prompt Engineering. The experimental results reveal the limitations of using general-purpose LLMs for intercepting low-quality code review comments.

First, \textbf{Aggressive Filtering.} Existing flagship models (such as Claude-4.5-Sonnet and Qwen-Coder-480B) demonstrate extremely high interception capabilities, e.g., Claude-4.5-Sonnet achieves an $\mbox{\it IR}$ of 71.0\% on the Github Testset. However, this comes at the cost of significant accuracy degradation—its $\mbox{\it FPR}$ reaches 46.4\%. This means the model blocks nearly half of the high-quality review comments. Similarly, the $\mbox{\it FPR}$ for DeepSeek-V3.2 and GLM-4.7 generally falls between 20\% and 40\%. This aggressive strategy leads to severe ``Information Loss" and is unsuitable for practical code review assistance scenarios.

Second, \textbf{Conservatism in Hybrid Reasoning (Thinking) Mode.} For models supporting hybrid reasoning, such as Qwen3-30B/235B, we observed an intriguing phenomenon: enabling the Thinking mode results in more conservative behavior. Compared to the No-think mode, the Thinking mode typically causes a simultaneous decrease in $\mbox{\it IR}$. For example, after enabling Thinking, the IR of Qwen3-30B drops from 34.6\% to 28.4\%. This suggests that while the Chain-of-Thought process increases reasoning depth, in subjective binary classification tasks like judging comment quality, the model tends to over-reason to find justifications for not intercepting, leading to false negatives.

Although our model has significantly fewer parameters than the aforementioned flagship models, thanks to high-quality domain-specific training data and a noise-resistant training framework, we outperform all LLM baselines according to the consolidated $\mbox{\it PIE}$ metric. 
We demonstrate that in the code review interception task, the \emph{Training Methodology} is more critical than mere \emph{Model Scaling}.

\vspace{-.2cm}
\subsection{Factor Analysis and Mechanism Study}
To understand the drivers behind \CRRM's performance, we analyze the effects of annotation volume, voting strategies, and the underlying data cleaning mechanism.
\vspace{-.3cm}
\paragraph{Impact of Annotation Volume.}
To verify \CRRM's practicality in cold-start scenarios, we default to using a minimal validator containing only 100 randomly sampled instances to guide the consensus cleaning process (denoted as \CRRM-MV-100). To further investigate the potential performance upper bound, we additionally annotated the dataset to reach 1,000 samples and compared the results. The comparison in Table~\ref{tab:codereview_refinement_performance} demonstrates the remarkable data efficiency of our method. Even with this extremely limited guidance, \textit{\CRRM-MV-100} significantly outperforms the standard fine-tuning baseline trained on noisy data. This confirms that a small amount of high-quality guidance is far more valuable than massive but noisy supervision. 

Expanding the annotation from 100 to 1,000 samples yields only marginal benefits.
While the full-data model \CRRM-MV-1k shows a slight increase in IR (Inner: 49.0\% $\rightarrow$ 56.6\%), the \CRRM-MV-100 model has already maintained an exceptionally low FPR (16.2\% on Github), matching the performance of the strictest full-data models. These findings suggest that \CRRM\ is highly robust. The initial investment of 100 annotations is sufficient to activate the purity-driven consensus-based data cleaning mechanism, while the substantial cost of expanding to 1,000 samples provides only a marginal utility increase. 
\vspace{-.3cm}
\paragraph{Impact of Voting Strategies.}
Based on the full-data model, we further analyze the impact of two pseudo-label aggregation strategies: SC and MV. The results reveal a clear trade-off suitable for different application scenarios:

First, \textbf{SC} exhibits strong conservatism. It tends to intercept only low-quality comments with extremely high confidence, thereby achieving the lowest $\mbox{\it FPR}$ among all methods (only 9.6\% on the Github Testset). Although its $\mbox{\it IR}$ is slightly lower, its exceptionally high $\mbox{\it PIE}$ makes it the optimal choice for scenarios with a near-zero tolerance for false positives. A Label Transition Matrix in Table~\ref{tab:transition_matrix} also confirms this phenomenon. The SC-100 strategy tends to reclassify a substantial number of samples originally labeled as ``Reject'' by users into the positive category, retaining only 30.23\% as negative. It systematically eliminates borderline or subjective negative samples, preserving only those indisputably low-quality. The underlying logic internalized by the model presumes a comment is valuable unless it exhibits glaring defects. This approach directly results in superior performance in terms of $\mbox{\it FPR}$ (Inner: 15.8 / Github: 13.8). The trade-off, however, is a slight reduction in $\mbox{\it IR}$, as the model lacks the confidence to intercept ambiguous cases.

Second, \textbf{MV} adopts a more balanced approach. Compared to SC, MV improves the $\mbox{\it IR}$ from 48.2\% to 56.6\% on the Inner dataset and from 42.4\% to 56.8\% on the Github dataset. Despite a slight increase in $\mbox{\it FPR}$, its overall performance remains robust and significantly superior to the unfiltered baseline. The MV strategy adopts a more noise-tolerant approach in exchange for broader coverage. As shown in Table~\ref{tab:transition_matrix}, MV-100 retains a significantly higher proportion of user rejections as negative (substantially exceeding that of SC). This percentage-point increment in negative sample retention represents the introduction of controversial or hard samples—comments that are disliked by some developers but deemed acceptable by others. Consequently, the training data constructed by MV exhibits higher diversity and variance within the negative class. Because the model is exposed to a wider spectrum of negative feedback (incorporating subjective rejection patterns), it learns a more generalized and aggressive decision boundary. The model becomes increasingly sensitive to potential issues, resulting in a higher $\mbox{\it IR}$ (Inner: 51.2 / Github: 51.8). However, this aggressiveness inevitably incurs the cost of over-generalization, leading to an increase in $\mbox{\it FPR}$ (rising from 13.8 to 16.2 on Github), as the model occasionally misclassifies reasonable but controversial comments as noise.

Last, both strategies also present similar patterns. First, the ``Ignore'' category 
is predominantly resolved as positive (approximately 83.4\% under SC and 83.98\% under MV). This indicates that 
silence often implies \emph{Tacit Acceptance} or \emph{Trivial Correctness} rather than \emph{Error}. Nevertheless, the method still identifies approximately 16.5\% of ignored comments as negative, likely accurately capturing the genuinely low-quality content or hallucinations that users completely disregarded. Second, for the ``Accept'' category, the model exhibits high stability, retaining over 95\% of the samples as positive. The marginal flip rate of approximately 4.5\% implies that while the model remains robust to user endorsements, it still performs a necessary degree of cleaning to correct for accidental acceptances.

This flexibility allows for strategic adjustments based on actual deployment requirements (i.e., prioritizing coverage versus prioritizing low interference). Comparing the two strategies, SC is more conservative than MV in assigning negative labels. SC flips a significantly higher proportion of ``Reject'' samples to positive (69.03\% vs. 62.51\%). By requiring a higher consensus threshold to mark a sample as noise, SC constructs a cleaner and safer training corpus. This directly explains why \CRRM-SC achieved the lowest $\mbox{\it FPR}$ (9.6\%), as it refuses to learn from ambiguous negative signals.

\vspace{-.2cm}
\begin{table}[htbp]
  \centering
  \scriptsize
  \caption{Label Transition Matrix: Original vs. Relabeled Labels}
  \label{tab:transition_matrix}
  \setlength{\tabcolsep}{4pt}
  \begin{tabular}{l c c c c}
    \toprule
    & \multicolumn{2}{c}{Strict Consensus (SC-100)} & \multicolumn{2}{c}{Majority Vote (MV-100)} \\
    \cmidrule(lr){2-3} \cmidrule(lr){4-5}
    \textbf{Origin} & Negative Label & Positive Label & Negative Label & Positive Label \\
    \midrule
    Accept & 976 (3.91\%)  & 24016 (96.09\%) & 1380 (5.52\%)  & 23612 (94.48\%) \\
    Reject & 7555 (30.23\%) & 17437 (69.77\%) & 11096 (44.40\%) & 13896 (55.60\%) \\
    Ignore & 3900 (15.60\%) & 21092 (84.40\%) & 5907 (23.64\%) & 19085 (76.36\%) \\
    \bottomrule
    \toprule
    & \multicolumn{2}{c}{Strict Consensus (SC-1k)} & \multicolumn{2}{c}{Majority Vote (MV-1k)} \\
    \cmidrule(lr){2-3} \cmidrule(lr){4-5}
    \textbf{Origin} & Negative Label & Positive Label & Negative Label & Positive Label \\
    \midrule
    Accept & 1093 (4.37\%)  & 23899 (95.63\%) & 1218 (4.87\%)  & 23774 (95.13\%) \\
    Reject & 7739 (30.97\%) & 17253 (69.03\%) & 9369 (37.49\%) & 15623 (62.51\%) \\
    Ignore & 4144 (16.58\%) & 20848 (83.42\%) & 4003 (16.02\%) & 20989 (83.98\%) \\
    \bottomrule
  \end{tabular}
\end{table}
\vspace{-.5cm}

\section{Conclusion}
Alert fatigue is a common challenge in applications where humans are assisted by intelligent agents such as LLM-based code review. This paper mitigates this issue by proposing a robust data filtering framework \CRRM\ to effectively learn from noisy user feedback and train a reflection model to intercept false alerts.

Unlike traditional noise-learning methods that rely on parameter initialization differences—which are often nullified in LLM fine-tuning—our approach creatively shifts the source of model divergence to the data space. By constructing a consensus mechanism across models trained on perturbed datasets with varying ambiguity ratios, we successfully extract effective training signals from high-noise user feedback under limited human supervision.

The resulting reflection model serves as a critical ``gatekeeper" at the final stage of intelligent agent pipelines such as code review assistants. Experimental results demonstrate that \CRRM\ significantly outperforms baselines trained on raw noisy data, achieving a superior balance between $\mbox{\it IR}$ and $\mbox{\it FPR}$, characterized by the consolidated metric - $\mbox{\it PIE}$. It effectively suppresses low-quality generation and hallucinations without silencing valuable insights, thereby directly mitigating alert fatigue. Beyond its immediate application as a filter, this framework also provides a high-quality Reward Model foundation for future Reinforcement Learning in many applications.

\section*{Impact Statements}
This paper presents work whose goal is to advance the pragmatic application of LLMs such as code review.

\paragraph{Impact on Deep Learning Research.}This paper introduces a novel data-driven methodology for fine-tuning model optimization directions in low-resource settings. By shifting model divergence from the parameter space to the data space through noise-doping and purity-driven co-teaching, we provide a flexible framework where developers can adjust the optimization bias—such as prioritizing precision over recall—simply by reconfiguring model combinations and consensus strategies. This approach minimizes the need for extensive human supervision and can be generalized to various domains, including scoring models, interception frameworks, and the construction of high-quality Reward Models for reinforcement learning. Furthermore, this research opens several promising theoretical avenues for future exploration, such as the formal characterization of convergence guarantees in purity-driven co-teaching and the development of theoretical bounds for error propagation in multi-model consensus under extreme noise. Investigating the information-theoretic relationship between ``consensus strength" and ``model generalization" remains a vital direction to further solidify the foundation of reflection modeling.

\paragraph{Mitigating Alert Fatigue in Human-AI Collaboration.} The pervasive deployment of LLMs in real-world applications, such as software development, has introduced a significant cognitive burden: ``Alert Fatigue." When AI-driven tools generate a high volume of low-quality or hallucinated feedback, human experts naturally become desensitized, leading to a ``crying wolf" effect where critical system vulnerabilities may be overlooked. This research provides a crucial ``purity-driven" filtering layer that restores trust in AI assistants. By ensuring that only high-signal, actionable insights reach the humans, our method fosters a more sustainable human-in-the-loop ecosystem. This shift from ``quantity-driven" to ``quality-driven" AI interaction reduces developer burnout and ensures that human cognitive resources are allocated to complex problem-solving rather than trivial noise verification.

\paragraph{Enhancing Software Safety and System Security.} Beyond individual productivity, the ability to accurately intercept low-quality code reviews has profound implications for global software safety. In mission-critical systems, an overlooked security flaw—buried under a mountain of AI-generated noise—can lead to catastrophic failures. Our framework’s emphasis on maintaining a near-zero False Positive Rate (FPR) while filtering ensures that the ``gatekeeping" process does not inadvertently suppress valid security warnings. Furthermore, the proposed consensus-based relabeling technique offers a blueprint for building more resilient automated safety auditors in other high-stakes domains, such as medical diagnostics or industrial monitoring, where the cost of a missed alert is equally high. By hardening the reliability of automated reflection models, this work contributes to the broader goal of deploying safe, trustworthy, and accountable AI systems in professional infrastructures.

\bibliography{cite}
\bibliographystyle{icml2026}



\end{document}